# Arabic Keyphrase Extraction using Linguistic knowledge and Machine Learning Techniques

**El-shishtawy T.A.**    &    **Al-sammak A.K.**

University of Benha
Faculty of Engineering (Shoubra)
shishtawy@hotmail.com

University of Benha
Faculty of Engineering (Shoubra)
sammaka@gmail.com

**Abstract**
In this paper, a supervised learning technique for extracting keyphrases of Arabic documents is presented. The extractor is supplied with linguistic knowledge to enhance its efficiency instead of relying only on statistical information such as term frequency and distance. During analysis, an annotated Arabic corpus is used to extract the required lexical features of the document words. The knowledge also includes syntactic rules based on part of speech tags and allowed word sequences to extract the candidate keyphrases. In this work, the abstract form of Arabic words is used instead of its stem form to represent the candidate terms. The Abstract form hides most of the inflections found in Arabic words. The paper introduces new features of keyphrases based on linguistic knowledge, to capture titles and subtitles of a document. A simple ANOVA test is used to evaluate the validity of selected features. Then, the learning model is built using the LDA - Linear Discriminant Analysis – and training documents. Although, the presented system is trained using documents in the IT domain, experiments carried out show that it has a significantly better performance than the existing Arabic extractor systems, where precision and recall values reach double their corresponding values in the other systems especially for lengthy and non-scientific articles.

## 1. Introduction

Keyphrases are list of phrases composed of about five to fifteen important words and phrases that express the main topics discussed in a given document. As the amount of electronic textual content grows fast, keyphrases can contribute to manage the process of handling these large amounts of textual information. Keyphrases play an important role in digital libraries, web contents, and content management systems, especially in cataloging and information retrieval purposes.

The limited number of documents that have author-assigned keyphrases as metadata description raises the need for a tool that can automatically extract keyphrases from text. Such a tool can enable many different types of information retrieval and analysis systems. It can provide the automation of:

- Generating metadata that gives a high-level description of a document's contents. This provides tools for text-mining related tasks such as document and Web page retrieval purposes.
- Summarizing documents for prospective readers. Keyphrases can represent a highly condensed summary of the document in question (Avanzo & Magnini, 2005).
- Highlighting important topics within the body of the text, to facilitate speed reading (skimming), which allows deciding whether it is relevant or not.
- Measuring the similarity between documents, making it possible to cluster and categorize documents (Karanikolas & Skourlas, 2006)**.**
- Searching: more precise upon using them as the basis for search indexes or as a way of browsing a collection of documents.

Many remarkable efforts have been proposed and implemented for automatically extracting keyphrases for English documents and other languages. In contrast, little efforts are achieved for documents written in Arabic language. Although, some researchers applied their keyphrase extraction systems to Arabic documents, but the proven efficiency of the extracted keyphrases was not satisfactory. This is the motivation for the present work, to propose a new methodology to enhance the performance of keyphrase extractor for Arabic documents.

In this paper, we introduce a methodology based on making use of both linguistic knowledge and machine learning techniques for extracting keyphrases of Arabic documents. Linguistic features and rules will be used during different steps of the extractor system. Two different machine learning models will be used to learn the system and classify candidate phrases as keyphrases or not. The results prove that the introduced methodology improves the accuracy of extracting keyphrases of Arabic documents relative to the available extractors.

## 2. Related Work

Several keyphrase extraction techniques have been proposed and implemented successfully in different context. Attempts on keyphrase extraction can be classified into two main streams, which are supervised machine learning algorithms and unsupervised machine learning algorithms. Most of the prior work in document keyphrases extraction problem is based on machine learning techniques. Turney (1997, 1999, 2000) approached the problem as a supervised learning task and presents two different machine learning algorithms for extracting keyphrases from a document. The first algorithm is based on the C4.5 decision tree classifier (Quinlan, 1993), and the second is the GenEx (Genitor and Extractor) algorithm (Turney, 1997, 1999, 2000). Turney's approach operates internally on stemmed phrases using the Lovins stemmer (Lovins, 1968) to learn a C4.5 decision tree classifier, and stemming by truncation for GenEx. A phrase here is treated as an n-gram of varying size (between 1 and 3 words) that does not contain any stop words. In both learning tasks (C4.5 and GenEx), 12 phrase parameters were identified, such as the number of words per phrase, the first occurrence of a phrase in a given document, the frequency of a phrase in an underlying document, etc. These parameters are used to encode the features of a given phrase within the input document. The C4.5 algorithm was trained on 9 of the 12 parameters, ignoring 2 features and using 1 feature as class prediction value. The input for the decision tree training was whole set of possible phrases, resulting in a very large proportion of feature vectors used as negative examples for keyphrases. The GenEx (Genitor and extractor) algorithm is used to reduce the amount of negative training examples. Extractor generates candidate phrases characterized by a dozen of numerical parameters. The setting of these parameters is determined by a training process, during which Genitor searches through the parameter space for values that yield a high overlap between the keyphrases assigned by the authors and the phrases that are output by extractor. After training, the best parameter values can be hard coded in extractor, and Genitor is no longer needed (Turney, 2002).

Kea (Frank et al., 1999; Witten et al., 1999, 2000) is another remarkable effort in this area, identifies candidate keyphrases in the same manner as Extractor. Kea then uses the Naïve Bayes algorithm to classify the candidate phrases as keyphrases or not. In Kea, candidate phrases are classified using only two features: (i) the TFxIDF, and (ii) the relative distance. The TFxIDF (term frequency times inverse document frequency) method which captures a word's frequency in a single document compared to its rarity in the whole document collection. It is used to assign a high value to a phrase that is relatively frequent in the input document (TF component), yet relatively rare in other documents (IDF component). The relative distance feature of a phrase in a given document is defined as the number of words that precede the first occurrence of the phrase divided by the number of words in the document. Kea uses the Naïve Bayes algorithm to calculate the probability of membership in a class (the probability that the candidate phrase is a keyphrase). Kea ranks each of the candidate phrases by the estimated probability that they belong to the keyphrase class. If the user requests N phrases, then Kea gives the top N phrases with the highest estimated probability as output.

KP-Miner (El-Beltagy & Rafea, 2008) is an unsupervised machine learning algorithm which uses the TFxIDF measures with two boosting factors. The first depends on phrase length, and the second depends on phrase position in the document. The KP-Miner system does not need to be trained on particular document set. It also has the advantage of being configurable, as the rules and heuristics adopted by the system are related to the general nature of documents and keyphrases. This implies that users can use their understanding of the input document to fine-tune the system to their particular needs.

## 3. The Proposed System

In this work, the automatic keyphrase extraction is treated as a supervised machine learning task. Two important issues are defined: how to define the candidate keyphrase terms, and what features of these terms are considered discriminative, i.e., how to represent the data, and consequently what is given as input to the learning algorithm. Our motivation is that adding linguistic knowledge (such as lexical features and syntactic rules) to the extraction process, rather than relying only on statistics, may obtain better results.

Thus, the current work is based on combining the linguistic knowledge and the machine learning techniques to extract keyphrases from Arabic documents with reasonable accuracy. The Linguistic knowledge will play important roles in different stages of our proposed system:

1. Analysis stage, where the document is tokenized into sentences and words. Each word is analyzed using an annotated Arabic corpus to extract its POS tags, category, and abstract form.
2. Candidate keyphrase extraction stage, where set of syntactic rules are used to determine the allowed sequence of words of the generated n-gram terms according to their POS tags and categories.
3. Features Vector calculation stage, where some of the selected features of each candidate phrase are linguistic-based, in addition to the statistical-based features.

The proposed system is based on three main steps: Linguistic pre-processing, candidate phrase extraction, and feature vector calculation. The following sections describe these steps in details.

## 4- Linguistic Pre-Processor

Each document must be preprocessed first to correct the writing mistakes as possible. Then, the input document is stemmed into sentences and words. Each word is analyzed to extract its lexical features like part of speech tags, category and abstract form. This process is described in the following subsections.

### 4.1. Document Preprocessing

The input document is segmented at two levels. In the first level, the document is segmented into its constituent sentences based on the Arabic phrases delimiter characters such as comma, semicolon, colon, hyphen, and dot. This process is useful for calculating part of the features vector of the candidate terms, such as Normalized Sentence Location (NSL), Normalized Phrase Location (NPL), Normalized Phrase Length (NPLen), and Sentence Contain Verb (SCV). In the second level, each sentence is segmented into its constituent words based on the criteria that words are usually separated by spaces. The generated sentences and words are exported to database tables related to their parent document table.

## 4.2. Part of Speech Analysis

The Analysis process concerns gathering linguistic knowledge of the input document. Complete document analysis requires morphology, syntax, and semantic analyses. This will burden the performance of the extractor, due to the heavy nature of these analyses. Thus, the focus will be on extracting the lexical features of the document words. Also, this type of analysis requires the presence of Arabic lexicon and Arabic morphological analyzer. Instead, we will make use of an annotated Arabic corpus to extract the required lexical features of the document words. Thus, a specially designed annotated Arabic corpus is used during the analysis process to enhance the speed of the proposed system. This corpus contains collections of analyzed Arabic words from different domains. The structure of the proposed corpus is defined by the word segments (prefix, stem, and suffix), word root, and abstract form. In addition to, the linguistic features of the word such as class, gender, count, and person. The most beneficial outputs from the analysis process are concluded by the word classification and abstract form. These features represent the basic linguistic knowledge required for enhancing the extraction process of candidate phrases and could be described as follows:

a) Word-Class: this attribute defines the classification of the given word as one of the possible Arabic word categories. In the present work, the word-class information is not shallow such as "noun", "verb", "adjective", or 'stop-word' only. Instead, the word-class is defined precisely to represent the linguistic usage of the given word within the context. Some examples of the word-class are: "general-noun", "count-noun", "place-noun", "time-noun", "proper-noun", "declined-noun", "augmented-noun", "adjective", "adverb", "past-verb", "present-verb", "ignore-verb", and more. The classification process makes use of other lexical parts of the given word like prefix, suffix, and stem. The prefix and suffix parts of the given word are used to refine the classical word classification. For example, the word "للبيانات" which contains the prefix part "لل", will be defined as "declined-noun اسم مجرور" instead of just "noun". The word "المرئية" contains the suffix part "ية" which will define the word-class as "adjective" and gender as "feminine". Also, the word "معلوماتهم" contains the suffix part "اتهم" which is used to define the word-class as "augmented-noun اسم مضاف", count as "plural", gender as "feminine", and person as "absent pronoun". This refinement is also valid even the word is a stop-word. For example, the word "إليهم" contains the suffix "هم" which is added to the word stem "إلى" that was originally classified as "preposition". The suffix part changes the word-class definition to be "declined-noun", "plural", and "masculine". This knowledge will be used in formulating the linguistic rules that control the process of extracting candidate key-terms and reduces the huge number of generated terms. In fact, this represents an efficient filter which minimizes the extracted key-terms to the lowest as possible.

b) Abstract form: it describes the basic form from which the given word is logically derived. Usually, this from differs from the word stem form which is obtained after removing the prefix and suffix parts of the word. For example, the stem of the word "المرئية" is "مرء", which represents a human being object. In contrast, the abstract form of the word is "مرئي", which represents the adjective of visual object. This abstract form can be used to represent many different words having the same logical meaning "Visual object" such as "النمذجة" , "النموذج المرئي" , "النماذج المرئية" , "المرئية". The abstract form of the given word is represented as follows:
- The single form for nouns.
- The single and male form for adjectives.
- The past form for verbs.
- The stem form for stop-words.

The abstract form of the given word is extremely useful during the process of extracting candidate keyphrases. For example, the words "المشروع" and "المشاريع" have different word-stems defined as "مشاريع" and "مشروع" respectively. But, their abstract forms are the same and defined as "مشروع". This abstract form is used for extracting candidate keyphrases by recommending a strong key-term like المشروع to represent the terms " مشروع الكتروني" "المشاريع الالكترونية" and "الالكتروني". This unified key-term can not be achieved by using the word-stem form of the words.

## 5. Candidate Phrases Extraction

In this phase, all possible phrases of one, two, or three consecutive words that appear in a given document are generated as n-gram terms following some syntactic rules. For example, the phrase "يقوم المدرس بشرح المناهج الدراسية" will produce a list of candidate phrases containing ("المدرس"، "المدرس بشرح"، "بشرح"، "بشرح المناهج"، "بشرح المناهج الدراسية"، "المناهج"، "المناهج الدراسية").
In the current work, a different approach is followed based upon the syntactic POS (Part of Speech tagging) analysis instead of relying only upon stop-word lists. In Arabic, stop-words are important and removing them lead to ignoring many useful structures. For example a preposition is not allowed to be in neither the first nor the last word in a keyphrase, but is allowed in any other position.
After investigating different keyphrases written for Arabic documents, we found that the following syntactic rules are effective for extracting candidate phrases:

1- The candidate phrase can start only with some sort of nouns like general-noun, place-noun, proper-noun, and declined-noun.
2- The candidate phrase can end only with general-noun, place-noun, proper-noun, declined-noun, time-noun, augmented-noun, adjective, and adverb.
3- For three words phrase, the second word is allowed to be count-noun, conjunction, preposition, and comparison, in addition to those cited in rule 2.

It is worthwhile to note that the used rules are language-dependant, and the given rules are applicable only to Arabic language. Table 1 shows an example of applying the extraction rules to an Arabic sentence. It is clear that the phrase "مشاريع التعليم عن" is excluded due to violation of rule 2. Also, the phrase "عن بعد" is excluded due to violation of Rule 1.

| Abstract form of (CP) | Candidate Phrases (CP) |
|---|---|
| Sentence: "إن مشاريع التعليم عن بعد تعتبر من أهم تقنيات الاتصالات والمعلومات." ||
| مشروع | مشاريع |
| مشروع تعليم | مشاريع التعليم |
| تعليم | التعليم |
| تعليم عن بعد | التعليم عن بعد |
| بعد | بعد |
| تقنية | تقنيات |
| تقنية اتصال | تقنيات الاتصالات |
| تقنية اتصال معلومة | تقنيات الاتصالات والمعلومات |
| اتصال | الاتصالات |
| اتصال معلومة | الاتصالات والمعلومات |

Table 1: Candidate phrases and their abstract forms

The final step of the candidate phrase extraction is to extract the abstract form of the candidate phrase words. Unlike English language, Arabic stem of the word (found by removing suffixes and prefixes) is not enough to decide that two words (and hence phrases) are similar. This is because the differences of Arabic word forms according to their count, gender and tense. The abstract form will be used to represent similar words. For example, the phrases (قواعد البيانات) and (قاعدة بيانات) are converted to (قاعدة بيان). This allows the algorithm to treat the two phrases as the same and generating a strong key-term. Examples of abstract forms of candidate phrases are shown in table 1. The abstract form of each word is extracted directly from the Arabic corpus[1].

## 6. Feature Vector Calculation

Each candidate phrase is assigned a number of features used to evaluate its importance. In our algorithm, three factors control the selection of features and their values.
1) The absolute importance of the phrase, which identifies its importance independent of its original document. Therefore, most feature values are normalized when necessary, to have ranges from zero to one.
2) Heuristics: where the feature values are computed, based on our hypothesis of its importance, after investigating many human written keyphrases.
3) All the extracted features and values are based upon the abstract form of the phrases.
The following features are adopted:
a) Normalized Phrase Words (NPW), which is the number of words in each phrase normalized to the maximum number of words in a phrase. The values of this feature can be 1, 1/2, or 1/3. The hypothesis is that keyphrases consists of three words are better than keyphrases contain two words, and so on..
b) The Phrase Relative Frequency (PRF), which represents the frequency of abstract form of the candidate phrase normalized by dividing it by the most frequent phrase in the given document. PRF has a maximum value of 1; when the candidate keyphrase is the most frequent one in a given document.
c) The Word Relative Frequency (WRF): The frequency of the most frequent single abstract word in a candidate phrase, normalized by dividing it by the maximum number of repetitions of all phrase words in a given document. The feature is calculated as follows: First, the frequency of all unique abstract words used in phrases for a given document is computed. Second, the maximum number of repletion is found, and used to normalize the computed frequencies. Third For each phrase, the maximum normalized frequency of its words is selected as a WRF. WRF has a maximum value of 1, when it contains the most frequent word of all words of phrases in a given document .
d) Normalized Sentence Location (NSL), which measures the location of the sentence containing the candidate phrase within the document. We use the heuristic that keyphrases located near the beginning and end of document are important phrases. We use the simple distribution function NSL= $(2(I/m)-1)^2$, where I is the location of the sentence within a document divided by total sentences in that document (m). The maximum value of NSL is 1 for first (I=0), and last sentences (I=m) in the document.
e) Normalized Phrase Location (NPL) feature is adopted to measure the location of the candidate phrase within its sentence. The NPL is given by $(2(x/n)-1)^2$, where x is the occurrence location of the phrase within a sentence divided by the total number of words of that sentence (n). Our motivation is that important keyphrases occur near the beginning and ending of sentences.
f) Normalized Phrase Length (NPLen), which is the length of the candidate phrase (in words), divided by the number of words of its sentence. This feature has a value of one, when the whole sentence is a keyphrase. Our hypothesis is that this will capture titles and subtitles of the document, which are likelihood to contain keyphrases .
g) Sentence Contain Verb (SCV). This feature has a value of zero if the sentence of the candidate phrase contains verb, else it has a value of one. Our motivation is that, this feature will give more weight to keyphrases written in titles and subtitles of a document. The feature value is assigned after analyzing the part of speech of sentence words.
h) Is It Question (IIT): This feature has a value of one if the sentence of the candidate phrase is written in a question form; else its value is 0. The hypothesis is that some authors highlight their main concepts as question forms. The feature is adopted to capture the important keyphrases written in documents as questions. During this work, question forms are only identified by part of speech tagging, when detecting question marks and/or question words.

---
[1] For example the Abstract form of the word (حمراء) is (أحمر), and for (سيأكل) is (أكل) and for (شجرة) is (أشجار).

i) (Is-Key): This feature is used only during the training phase. It has a value of one if the candidate phrase matches one of the author-assigned keyphrases.

However, many authors; starting from Turney (1997, 1999, 2000), used the features (a), (b) and (c), the proposed algorithm uses different normalization technique to satisfy our hypothesis of feature importance. Also, feature (i), normally used in supervised learning algorithms. Finally, the original form of candidate abstract keyphrase form is retained for presentation to the user in case the phrase does turn out to be a keyphrase. This process is a straightforward operation. The proposed algorithm is computed for all candidates instead of unique stemmed keyphrases (KEA and Turney), which eliminates the need for selecting the most frequent keyphrase, when several different forms occur.

## 7. Learning Experiments

In the previous section, we propose eight statistical and linguistic features to represent the importance of each candidate phrase. Now, the degree of effect of each feature during classification of candidate phrases as keyphrases or not will be measured. This objective can be achieved by using learning model and training documents having author-assigned keyphrases.
ANOVA (ANalysis Of VAriance) test is used to evaluate the validity of the selected features. The learning model is then built using the LDA (Linear Discriminant Analysis) and training documents with known key phrases. Finally the model is used to find key phrases for new documents. The following subsections reviews the size of the training data, ANOVA test, and LDA model.

### 7.1 Training Data

A sample of 30 documents is collected, preprocessed, analyzed, and manually keyphrase-assigned. The training data are collected from different sources and domains with focus on the computer area. It includes journal articles, technical reports, papers, and sections from textbooks. Little of these documents have author-assigned keyphrases. The other documents are read carefully by many specialists in the field to assign the proper document keyphrases. These data will be used to train the proposed system by adjusting the classifier parameters for best match with assigned keyphrases.
The training sample produces 6671 candidate keyphrases. The author's assigned keyphrases, which ranges from 5 to 7 keyphrases per document, are mapped to the Is-Key feature. Each occurrence of the abstract form of these keyphrases are marked, which gives 885 positive and 5786 negative keyphrases.

### 7.2 ANOVA for Regression

The analysis of variance (ANOVA) for regression test can be used to measure the effect of changes in parameter values on the dependant output. In our case, the "feature vector" for candidate phrases will represent the input effective parameters, and the "Is-Key" feature represents the dependant output. Table 2 describes the effects of the eight parameters of the feature vector (x1, x2, …x8) on the "Is-Key" output. The experiment proves that the parameter x5 (PRF) is the most effective one, whereas the parameter x8 (IIT) is the least effective one, and the parameters x3 (NPL), x7 (SCV) have no effect on the output.

| Feature (input parameter) | Effect ($R^2$) |
|---|---|
| X1 (NPW) | 0.004 |
| X2 (NPLen) | 0.005 |
| X3 (NPL) | 0.000 |
| X4 (NSL) | 0.003 |
| X5 (PRF) | 0.116 |
| X6 (WRF) | 0.039 |
| X7 (SCV) | 0.000 |
| X8 (IIT) | 0.001 |

Table 2: The effect of feature vector on the "Is-Key" output

Also, table 3 shows the accumulative effect of the input parameters on the output. It is clear that the effect on the output increases as the number of accumulated affecting parameters increases. The last row in the table proves that the highest impact on the output is achieved by accumulating the effects of all effective parameters x5, x6, x2, x1, x4, and x8.

| Model | Accumulated Effect ($R^2$) |
|---|---|
| X5 | 0.116 |
| X5, x6 | 0.155 |
| X5, x6, x2 | 0.159 |
| X5, x6, x2, x1 | 0.164 |
| X5, x6, x2, x1, x4 | 0.167 |
| X5, x6, x2, x1, x4, x8 | 0.168 |

Table 3: The accumulated effect of effective features on the "Is-Key" output

### 7.3 Linear Discriminant Analysis (LDA)

In the machine learning approach, the goal of classification is to group candidate keyphrases that have similar feature values, into two groups. The keys in first group are classified as "yes", while the others are classified as "no". A linear classifier achieves this by making a classification decision based on the value of the linear combination of the features. If the input feature vector to the classifier is a real vector $\vec{x}$, then the output score is :

$$y = f(\vec{w} \cdot \vec{x}) = f\left(\sum_j w_j x_j\right)$$

Where, $\vec{w}$ is a real vector of weights and $f$ is a function that converts the dot product of the two vectors into the desired output. The weight vector $\vec{w}$ is learned from a set of labeled training samples. The function $f$ maps all values above a certain threshold to the first class (keyphrase) and all other values to the second class (not a keyphrase). The classification is based on Bayes rule, which assigns an object to the group with highest conditional probability. The Linear Discriminant Analysis formula which assigns a keyphrase to group i that has maximum $f_i$ is given by:

$$f_i = \mu_i c^{-1} x_i^T - \frac{1}{2} \mu_i c^{-1} \mu_i^T + Ln(p_i)$$

Where **P** is the prior probability about group **i**, μ is the mean of features in group i and **C** is the covariance matrix of group **i**.

## 8. Evaluating Results

In order to evaluate the performance of the proposed system, three experiments were carried out to test the proposed system. Three data sets containing a total of 50 documents were used. The first experiment aimed to measure the level of acceptance of the extracted keyphrases. Since there is no author-assigned keyphrases for these documents, a human judge was adopted to evaluate this level (Barker & Cornacchia, 2000). The judges in our experiment are university faculty and postgraduate students.

In second and third experiments, keyphrases extracted by the presented system were compared to those extracted by two Arabic keyphrase extraction systems: KP-Miner (web link http://www.claes.sci.eg/coe_wm/kpminer), and Sakhr Keyword Extractor (web link http://www.sakhr.com/ Technology/ Keyword/ Default.aspx? sec=Technology &item= KeywordS). The reason for choosing KP-Miner and Sakhr is that both systems have available executable versions on the web, and hence can be tested on the same documents. Second experiment evaluates the presented system in Information Technology domain, which is the same domain used to train our system. To investigate how our system performs with non-scientific writings, experiment 3 was carried out on the social domain.

In experiments 2 and 3, Precision (P) and Recall (R) metrics are used to evaluate the performance of the proposed extractor. Precision is an estimate of the probability that a given model classifies a keyphrase as relevant to a user's keyphrases. Recall is an estimate of the probability that, if a keyphrase is relevant to a user's keyphrases, then a given model will classify it as relevant.
**P** is given by **a/(a+b)**, and **R** is given by **a/(a+c)**, where:
**a**: keyphrases classified by both of the model and human.
**b**: keyphrases classified by the model, but classified as not keyphrases by human.
**c**: keyphrases classified as not keyphrases by model, but classified as keyphrases by human.
The details of each experiment are given below.

### 8.1 Experiment 1

In the first experiment, a dataset of 10 documents containing short news about the Egyptian ministry of Information and Communication was used. The average length of documents is 203 words. Each judge is given a list of twelve keyphrases extracted by our system together with the original document. Judges rated each keyphrase as accepted (1) or not (0). The score for each keyphrase was calculated simply as the sum of the scores from all judges. A keyphrase accepted, if it got more than 50% of the votes. The process was then repeated for all documents in the dataset to compute the overall average of acceptance. Table 4 summarizes the experiment features. Table 5 shows an example of the extracted keyphrases for one document of the dataset using our proposed extractor and the KP-Miner extractor. The accepted keyphrases by a human judge is underlined and bolded. The results show that, averages of 44% of the extracted keyphrases by our system were accepted by human judges.

| Total No of Documents | 12 |
|---|---|
| Average words/Document | 203 |
| Extracted keyphrases /Document | 12 |
| Average of acceptance | 44.16% |
| Standard deviation of acceptance | 6% |

Table 4: Summary of experiment 1

| | |
|---|---|
| **Our System** | |
| **Kp-Miner** | |

Table 5: Example of extracted keyphrases for a document in experiment 1 dataset

### 8.2 Experiment 2

In this experiment, keyphrases extracted by the presented system were compared to those extracted by KP-Miner and Sakhr Extractors. A dataset of a randomly collected 20 documents in IT domain was used to test the three systems. The average size of documents is 376.1 words per document. Table 6 shows the precision and recall values for five, seven and ten extracted keyphrases of the three systems. The results prove that the presented system performs slightly better than the KP-Miner, and both outperform Sakhr system in terms of precision and recall.

| # of Key phrases | Sakhr | | KP- Miner | | Our System | |
|---|---|---|---|---|---|---|
| | P | R | P | R | P | R |
| 10 | 0.20 | 0.14 | 0.48 | 0.30 | **0.53** | **0.36** |
| 7 | 0.20 | 0.17 | 0.39 | 0.34 | **0.52** | **0.46** |
| 5 | 0.17 | 0.20 | 0.36 | 0.44 | **0.44** | **0.56** |

Table 6: Precision (P) and Recall (R) for experiment 2 dataset

Table 7 shows a sample of the extracted keyphrases of the three systems, with those matching author keyphrases are underlined and bolded. .

| Our System | |
|---|---|
| Kp-Miner | Relational |
| Sakhr | |

Table 7: An example of extracted keyphrases in the IT domain by the three systems

### 8.3 Experiment 3

In this experiment, a dataset of 20 documents with an average of 675.2 words/document containing non-scientific topics is collected from newspapers and Arabic web sites. Table 8 shows the precision and recall values for five, seven and ten extracted keyphrases of the three systems. It is clear from the results that the proposed system outperforms both of KP-Miner and Sakhr systems in non-scientific and lengthy documents. The precision of our system ranges from 1.9 to 1.3 that of KP-Miner and the recall ranges from 2 to 1.3 the corresponding values of KP-Miner. By comparing the results shown in tables 6 and 8, it is clear that the precision of the KP-Miner system decreases from 0.48 to 0.34 (for ten extracted keyphrases) when tested using non-scientific documents. This corresponds to an increase from 0.53 to 0.65 for our system. This comes from the existence of more inflected words in non-scientific writings, whose similarities are easily detected by our system through the use of the abstract form of words. Table 9 shows an example of the extracted keyphrases from the three systems for one of the dataset documents, with those matching author keyphrases are underlined and bolded.

| # of Key phrases | Sakhr | | KP-Miner | | Our System | |
|---|---|---|---|---|---|---|
| | P | R | P | R | P | R |
| 10 | 0.19 | 0.12 | 0.34 | 0.20 | **0.65** | **0.40** |
| 7 | 0.14 | 0.11 | 0.30 | 0.26 | **0.53** | **0.46** |
| 5 | 0.10 | 0.12 | 0.27 | 0.32 | **0.35** | **0.44** |

Table 8: Precision (P) and Recall (R) for the Social domain documents

| Our System | |
|---|---|
| Kp-Miner | |
| Sakhr | |

Table 9: An example of extracted keyphrases in non-scientific (social) domain by the three systems

## 9. Conclusion and Future Work

In this paper we have shown how keyphrases extraction can be achieved by using statistical measures as well as linguistic knowledge from Arabic documents, as input to a machine learning algorithm. The main point of this paper is that by adding linguistic knowledge to the representation, a better result is obtained. Experiments carried out in this paper show that the presented system has a significantly better performance than the existing Arabic extractor systems, where precision and recall reach double their values in lengthy and non-scientific articles. One reason is that, using more linguistic knowledge allows efficient use of the traditional statistical measures. For example, in our work, the use of the abstract Arabic form of a word enhances all frequency-based statistical features, since it captures all words and phrases inflections. Second, there is a great role of the features 'Normalized Phrase Length' and 'Sentence Contain Verb', which capture most important keyphrases in titles and subtitles in lengthy documents. These features can not be calculated without syntactic knowledge POS tags. Third, the use of linguistic rules, which fed only candidate keyphrases that are reasonable from a human perspective to the learner.

Future works include many points such as the use of simpler compact corpus with light stemmer, instead of the sophisticated corpus used during the analysis phase. Second point is the effect of word class ambiguity on the system performance. Also, the use of enhancement filters such as sub-phrase and common-word removal.